\documentclass[conference]{IEEEtran}
\IEEEoverridecommandlockouts
\usepackage{amsmath,amssymb,amsfonts}
\usepackage{algorithmic}
\usepackage{graphicx}
\usepackage{textcomp}
\usepackage{xcolor}
\def\BibTeX{{\rm B\kern-.05em{\sc i\kern-.025em b}\kern-.08em
    T\kern-.1667em\lower.7ex\hbox{E}\kern-.125emX}}

\usepackage[utf8]{inputenc} % allow utf-8 input
\usepackage[T1]{fontenc}    % use 8-bit T1 fonts
\usepackage{hyperref}       % hyperlinks
\usepackage{url}            % simple URL typesetting
\usepackage{booktabs}       % professional-quality tables
\usepackage{amsfonts}       % blackboard math symbols
\usepackage{nicefrac}       % compact symbols for 1/2, etc.
\usepackage{microtype}      % microtypography
\usepackage{xcolor}         % colors
\usepackage{graphicx}       % for including graphics
\usepackage{float}
\usepackage{placeins}
\usepackage{tcolorbox}  % For creating boxes
\usepackage{amsmath}    % For math formatting
\usepackage{subfigure}
\usepackage{cite}

\begin{document}

\title{Can a Large Language Model Learn Matrix Functions In Context?\\}

\author{\IEEEauthorblockN{Paimon Goulart}
\IEEEauthorblockA{\textit{Computer Science and Engineering} \\
\textit{University of California Riverside}\\
Riverside, CA, USA \\
paimon.goulart@email.ucr.edu}
\and
\IEEEauthorblockN{Evangelos E. Papalexakis}
\IEEEauthorblockA{\textit{Computer Science and Engineering} \\
\textit{University of California Riverside}\\
Riverside, CA, USA \\
epapalex@cs.ucr.edu}
}

\maketitle

\begin{abstract}
Large Language Models (LLMs) have demonstrated the ability to solve complex tasks through In-Context Learning (ICL), where models learn from a few input-output pairs without explicit fine-tuning \cite{min2022rethinking}. In this paper, we explore the capacity of LLMs to solve non-linear numerical computations, with specific emphasis on functions of the Singular Value Decomposition. 
%including vector P-Norms, Nuclear Norms, and the Top-K Singular Values of matrices. 
Our experiments show that while LLMs perform comparably to traditional models such as Stochastic Gradient Descent (SGD) based Linear Regression and Neural Networks (NN) for simpler tasks, they outperform these models on more complex tasks, particularly in the case of top-k singular values. Furthermore, LLMs demonstrate strong scalability, maintaining high accuracy even as the matrix size increases. Additionally, we found that LLMs can achieve high accuracy with minimal prior examples, converging quickly and avoiding the overfitting seen in classical models. These results suggest that LLMs could provide an efficient alternative to classical methods for solving high-dimensional problems. Future work will focus on extending these findings to larger matrices and more complex matrix operations while exploring the effect of using different numerical representations in ICL.\end{abstract}

\section{Introduction}
In-Context Learning (ICL) is an inference technique performed on Large Language Models (LLMs) that enables them to solve problems previously thought to require task specific training. By conditioning the model on varying numbers of input-label pairs \cite{min2022rethinking}\cite{wei2022chain}, LLMs can perform a wide range of tasks—like regression \cite{coda2023meta},  graph computations \cite{fatemi2023talk}, and knowledge graph completion \cite{lee2023temporal} —without the need for fine-tuning. This showcases the versatility of LLMs, allowing them to generalize across different tasks while minimizing the traditional training overhead, making them suitable for more dynamic applications. 

In this paper, we explore the ability of LLMs to learn various vector and matrix functions in context, focusing on tasks like the p-norm, nuclear norm, and top-k singular values of a matrix. While it is already known that In-Context Learning (ICL) can effectively handle linear regression tasks \cite{coda2023meta}, we aim to investigate how well it performs on more complex, non-linear tasks. These problems, while solvable by deterministic algorithms with perfect accuracy, scale poorly as the size of the target matrix increases \cite{golub2013matrix}. By applying ICL, we hypothesize that LLMs can provide a faster alternative for approximating solutions. With a handful of demonstrations, we expect the LLM to learn the patterns of these non-linear tasks and predict accurate solutions, even within large and complex problem spaces. If successful, this approach could significantly reduce computational overhead and enable the efficient solving of high-dimensional, non-linear problems.

From our experiments, we investigated how well an LLM performs at computing the p-norm of vectors, as well as the nuclear norm and top-k singular values of a matrix.

\section{Proposed Method and Experimental Setup}

\subsection{Tested Functions}

\paragraph{P-norm of Vectors}
The p-norm, also known as the $\ell_p$ norm, is a generalization of vector magnitude in a finite-dimensional space \cite{horn_matrix_1985}. It measures the length of a vector according to a specified value of \( p \).
The p-norm of a vector \( \mathbf{x} \in \mathbb{R}^n \) is given by:

\[
\|\mathbf{x}\|_p = \left( \sum_{i=1}^{n} |x_i|^p \right)^{1/p}
\]

\paragraph{Top-k Singular Values of a Matrix}

The top-k singular values of a matrix refer to the largest \( k \) singular values obtained from the singular value decomposition (SVD) \cite{golub2013matrix}.

For an arbitrary matrix, its singular values can be obtained by finding a set of \( A \in \mathbb{R}^{m \times n} \) matrix decomposition satisfying the following form: 
$
A = U \Sigma V^T
$

where:
\begin{itemize}
    \item \( U \in \mathbb{R}^{m \times m} \) is an orthogonal matrix containing the left singular vectors.
    \item \( \Sigma \in \mathbb{R}^{m \times n} \) is a diagonal matrix containing the singular values \( \sigma_1 \geq \sigma_2 \geq \dots \geq \sigma_{\min(m, n)} \) on its diagonal.
    \item \( V \in \mathbb{R}^{n \times n} \) is an orthogonal matrix containing the right singular vectors.
\end{itemize}

The top-k singular values are the first \( k \) entries in the diagonal of \( \Sigma \) and are uniquely defined for any arbitrary matrix.

\paragraph{Nuclear Norm of Matrices}

The nuclear norm is used to measure the sum of the singular values of a matrix \cite{recht2010guaranteed}. 

For a matrix \( A \in \mathbb{R}^{m \times n} \), the nuclear norm is defined as the sum of the singular values \( \sigma_i \) of the matrix:

\[
\|A\|_* = \sum_{i=1}^{\min(m, n)} \sigma_i
\]

where \( \sigma_i \) are the singular values obtained through singular value decomposition (SVD) of the matrix \( A \).

\subsection{Method formulation and prompting setup}

As shown in \autoref{fig:example_prompt}, the prompt (inspired by meta in-context learning \cite{coda2023meta} and the provided code base\footnote{https://github.com/juliancodaforno/meta-in-context-learning}) is set up in such a way that no matter what question we decide to ask it, we do not have to adjust the prompt in any way. Whether it is a norm with a scalar output or the top-k singular values with a vector output, we can use the same prompt for both tasks. This is vital in ensuring that we are not introducing any bias when asking different types of questions. 

Another important point is that we are never explicitly nor directly giving the task to the LLM. In other words, we never tell it whether it is solving a vector norm, a matrix norm, or solving the top-k singular values of a matrix. We just give it an input and an output and ask it to predict a new question based on what it observed previously. Essentially, we are setting up the problem as a non-linear regression task for the LLM. We want to see if, given enough training examples, the LLM can recognize this task as regression and fit some sort of function over the data.

An additional quality of our setup is its transferability. The same prompt structure can be easily applied to a variety of downstream tasks. Since we also want to observe how an LLM performs when compared against more conventional methods such as Stochastic Gradient Descent (SGD), a 2-layer Neural Network (NN), or a Convolutional Neural Network (CNN), the setup of our problem ensures that the LLM is not given any advantage or disadvantage from how it is prompted.

\begin{figure}[!htp]
    \centering
    \begin{tcolorbox}[colframe=blue!50!black, colback=blue!5!white, title=Example Prompt]
        You observe a machine that produces an output $y$ for a given input $x$:
        \vspace{0.3cm}
        
        Machine 1:
        
        If no previous examples are available, sample $y$ from your prior distribution. Provide only the output, formatted as a set of values that matches the length of the previous $y$'s, followed by semicolons. No words, only numbers and semicolons. Even if you are unsure, try to find a pattern and predict $y$ as accurately as possible.

        \vspace{0.3cm}

        Examples:
        \[
        \begin{aligned}
        \quad x &= [[n_{00}, n_{01}, n_{02}, n_{03}, n_{04}], [n_{10}, n_{11}, n_{12}, n_{13}, n_{14}], \\
           &\phantom{[} [n_{20}, n_{21}, n_{22}, n_{23}, n_{24}], [n_{30}, n_{31}, n_{32}, n_{33}, n_{34}], \\
           &\phantom{[} [n_{40}, n_{41}, n_{42}, n_{43}, n_{44}]] \\
        \quad y &= \lambda_{1}; \lambda_{2}; \ldots; \lambda_{n};
        \end{aligned}
        \]
        \vspace{-0.3cm}
        \begin{flushleft}
        \quad $\cdots$ (Varying number of examples); \\
        \quad Given $x = (\text{Input vector or matrix})$, predict $y$:\_\_\_\_; 
        \end{flushleft}

    \end{tcolorbox}
    \caption{Example prompt with 5x5 matrix, input can also be set as a vector or a matrix of any size. Example outputs denoted as $\lambda$ represent the output as either a singular scalar, or vector (adapted from Coda-Forno et al. \cite{coda2023meta}.)}
    \label{fig:example_prompt}
\end{figure}

\subsection{Baselines for Comparison} 
As mentioned earlier, the baselines for our comparisons will be drawn from the RMSE of SGD Linear Regression (SGD), a 2-layer Neural Network (NN) \cite{xu2022sv}, and a Convolutional Neural Network (CNN) when dealing with matrix operations.

We chose these models as baselines because they are commonly used to solve regression problems and are known for their effectiveness in their respective domains. By using these models, we can evaluate the performance of the Large Language Model (LLM) in comparison to more traditional methods. If the LLM performs as well as, or better than, these models, it can demonstrate the potential for replacing or complementing classical methods in specific scenarios.

\section{Experiment Procedure}

For our experiments, we randomly generated a sequence of vectors $(A_i)_{i=0}^N$, where $A_i$ is either an \(n\)-dimensional vector or an \(n \times n\) matrix with entries drawn from the real number range \([-100, 100]\). We then employed NumPy \cite{harris2020array} to compute the corresponding \(p\)-norm, nuclear norm, or the top three singular values of the matrix. For our experiments we used 50 examples in total ($N = 50$).

For each of the target quantities, we began evaluation with one prior example. The first matrix, along with its computed solution, was used as the prior example, while the second matrix was presented as an input with its solution withheld. Following this, we then evaluate the RMSE of the model's prediction for the held out dataset with that of the ground truth. We repeat this for each matrix $A_i, i \in \{ 2 \cdots N \}$, where the previous examples $\{ A_{j} \}_{j<i}$ are used for training. The model is evaluated on $A_i$.

We employed the prompt setup described in \autoref{fig:example_prompt} to present these examples to the selected LLM. If the LLM's output did not adhere to the expected format, the query was regenerated until the output format was correct. Once a correctly formatted response was obtained, we computed the Root Mean Square Error (RMSE) between the LLM’s output and the actual precomputed solution.

For the comparison models, including Stochastic Gradient Descent (SGD)-based Linear Regression, a 2-Layer Neural Network (NN), and a Convolutional Neural Network (CNN), we trained each model on the same set of prior examples provided to the LLM. These models were all trained to convergence. 

To select optimal hyperparameters, such as learning rate, we conducted a grid search \cite{bergstra2012random}. For the learning rate, we tested values of \(10^{-2}\), \(10^{-3}\), and \(10^{-4}\). A holdout set was used to assess model performance across these configurations, and the hyperparameters yielding the best performance were selected. 

These models were subsequently tested on the same final matrix used in the LLM evaluation. We then calculated the RMSE between the predicted values from these models and the actual computed values.

All our code containing the experimental setup as well as plot generation can be found here: \url{https://github.com/Pie115/Learning-Matrix-Functions-In-Context/tree/main}.

\subsection{LLMs Used}
In our experiments, we mostly used the Gemini model \cite{team2023gemini}. Specifically, we utilized version `gemini-1.5-flash`. The model interacts with the prompt and returns the generated response through the Gemini API.

Along with this, we also wanted to see how other open source models would compare to the results of Gemini. To do this, we used both Qwen2.5-72B-Instruct \cite{qwen2}\cite{qwen2.5} and Hermes-3-Llama-3.1-8B \cite{teknium2024hermes3technicalreport}.
\section{Results \& Discussion}

In this section we demonstrate the performance in $L_p$ norms, which are simpler to compute, thus potentially more easily learnable. We also show more challenging cases, such as the functions of the singular value decomposition, nuclear norm of a matrix (which is equal to the sum of its singular values) and the top-3 singular values. 

\subsection{Learning Vector Norms In Context}

\begin{figure*}[!ht]
    \centering
    \includegraphics[width=0.9\textwidth, keepaspectratio]{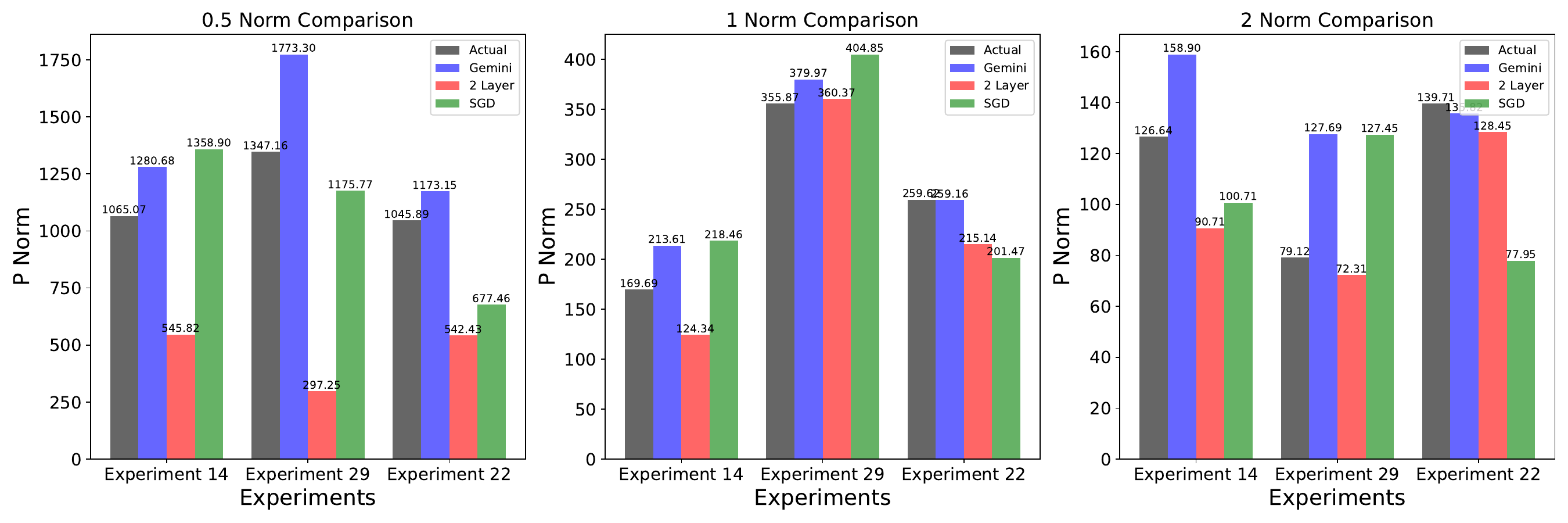}
    \caption{Average predictions compared to average actual values for the vector norm learning task.}
    \label{fig:P-Norm_average_predict}
\end{figure*}

\begin{figure*}[!ht]
    \centering
    \includegraphics[width=0.9\textwidth, keepaspectratio]{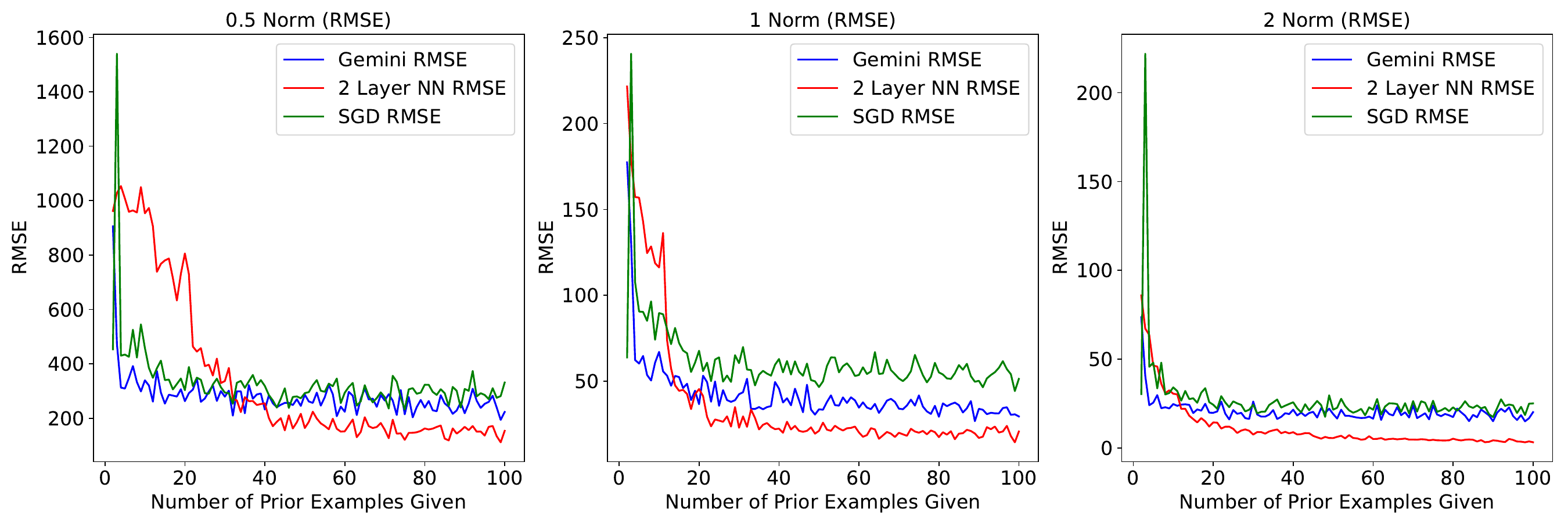}
    \caption{RMSE of the vector norm predictions for every model in the vector norm learning task.}
    \label{fig:rmse_P-Norm_all_models}
\end{figure*}

The figures below (\autoref{fig:P-Norm_average_predict} and \autoref{fig:rmse_P-Norm_all_models}) demonstrate the accuracy and error metrics of different models when predicting various vector norms.

Figure \ref{fig:P-Norm_average_predict} is divided into three subplots: Figures 1a, 1b, and 1c, which represent the predictions for the $p$-norm when $p = 0.5$, $p = 1$, and $p = 2$, respectively. Each bar plot shows the predicted vs actual values from three randomly selected experiments across all models. As seen in the bar plots, Gemini consistently performs either more accurately or on par with more classical models in these tasks.

To further illustrate these results, Figure \ref{fig:rmse_P-Norm_all_models} presents the Root Mean Square Error (RMSE) of each model's predictions. Like the previous figure, it is divided into three subplots: Figures 2a, 2b, and 2c, which correspond to $p = 0.5$, $p = 1$, and $p = 2$, respectively. Models with lower RMSE values demonstrate better performance in predicting the vector norms. As shown, before reaching approximately 20 prior examples, Gemini outperforms both the two-layer Neural Network (NN) and Stochastic Gradient Descent (SGD) Linear Regression models. After this threshold, the two-layer NN begins to slightly outperform Gemini, though the difference is minimal.

However, an interesting trend begins to emerge here: as the complexity of the task increases, Gemini tends to produce superior results more consistently. This trend suggests that Gemini may have an advantage in handling more complex functions and tasks—a hypothesis that will become more evident in the SVD-based tasks.
%in the following sections on nuclear norms and SVDs.

In summary, while Gemini demonstrates strong performance with fewer training examples, conventional approaches such as the two-layer Neural Network may offer a slight advantage in solving simpler tasks with more training data. 

\subsection{Learning Matrix Norms In Context}

\begin{figure*}[!ht]
    \centering
    \includegraphics[width=0.9\textwidth, keepaspectratio]{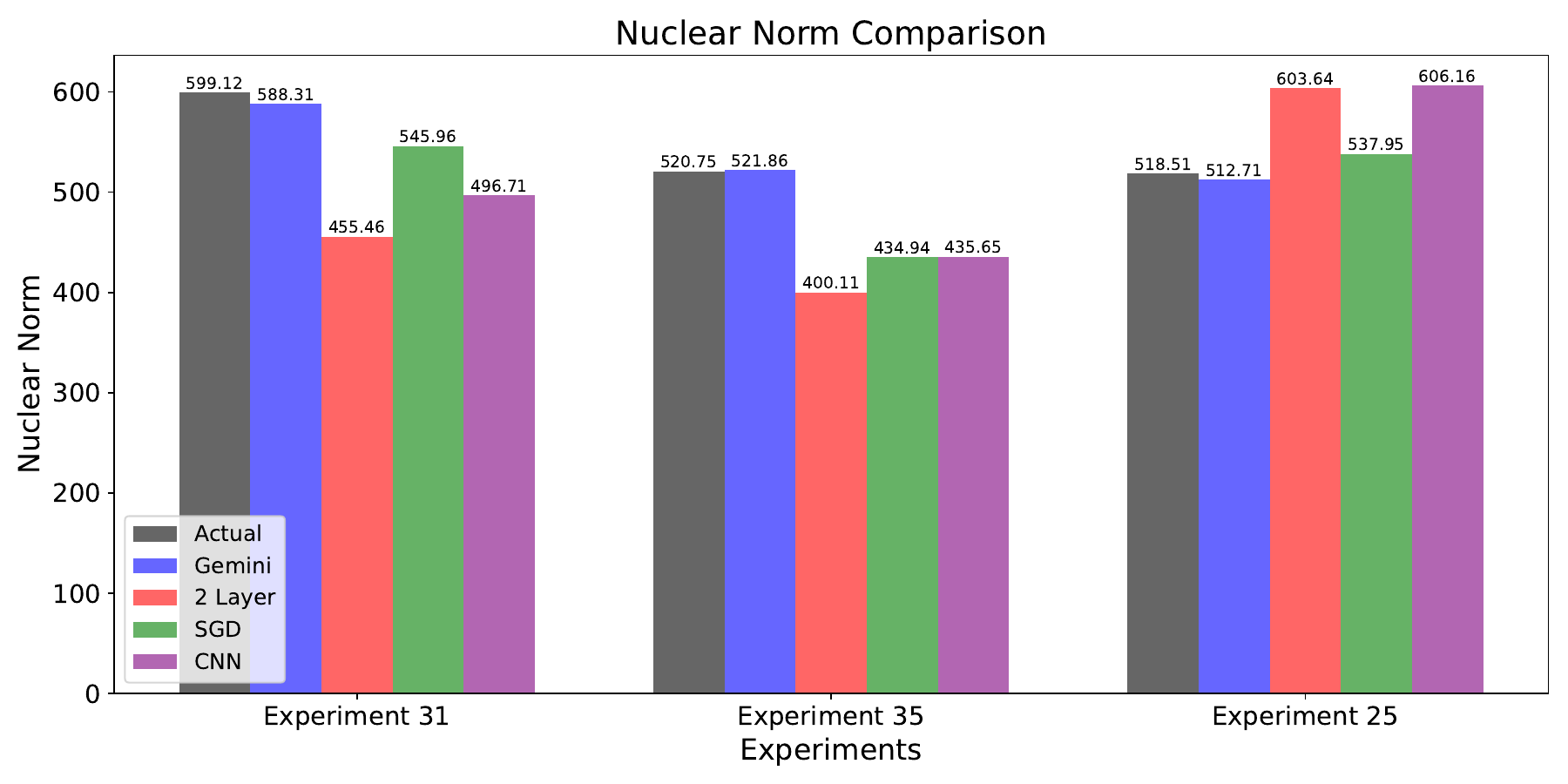}
    \caption{Average predictions compared to average actual values for the nuclear norm learning task at randomly selected experiments for 5x5 matrix inputs.}
    \vspace{1cm}
    \label{fig:nuclear_norm_average_predict}
\end{figure*}

The figures (\autoref{fig:nuclear_norm_average_predict} and \autoref{fig:rmse_nuclear_norm_all_models}) illustrate the performance of different models in predicting the nuclear norm of a randomly generated 5x5 matrix.

Figure \ref{fig:nuclear_norm_average_predict} presents a bar chart that compares the predicted vs actual values from 3 randomly selected experiments across all models. In these experiments, Gemini consistently outperforms the other models, demonstrating significantly more accurate predictions across all cases.

To further support this, Figure \ref{fig:rmse_nuclear_norm_all_models} shows the Root Mean Square Error (RMSE) of each model's predictions as a function of the number of prior examples. Gemini maintains the lowest RMSE, indicating superior accuracy compared to other models, particularly with fewer prior examples. Gemini's performance plateaus after around 20 prior demonstrations, suggesting that it requires fewer training examples to reach optimal accuracy. Despite this, Gemini continues to outperform classical models, including SGD Linear Regression, even with larger numbers of examples, showing strong generalization without overfitting.

These results indicate that Gemini not only achieves better results than classical models for the nuclear norm task but also does so with fewer training examples, highlighting its efficiency and robustness for this specific task. As we observed earlier with the $p$-norms, Gemini performs well even with simpler tasks. However, with the nuclear norm—a much more complex task than the $p$-norms—Gemini clearly begins to outshine the other models, reinforcing the trend that performance truly excels as task complexity increases.

\subsection{Learning Singular Values In Context}

\begin{figure*}[!ht]
    \centering
    \subfigure[Plot of RMSE of the nuclear norm predictions for every model in the nuclear norm learning task over all 100 prior examples. The table below shows the average RMSE of the nuclear norm over the last 25 trials. \label{fig:rmse_nuclear_norm_all_models}]{
        \begin{minipage}{0.45\textwidth}
            \centering
            \includegraphics[width=\textwidth, keepaspectratio]{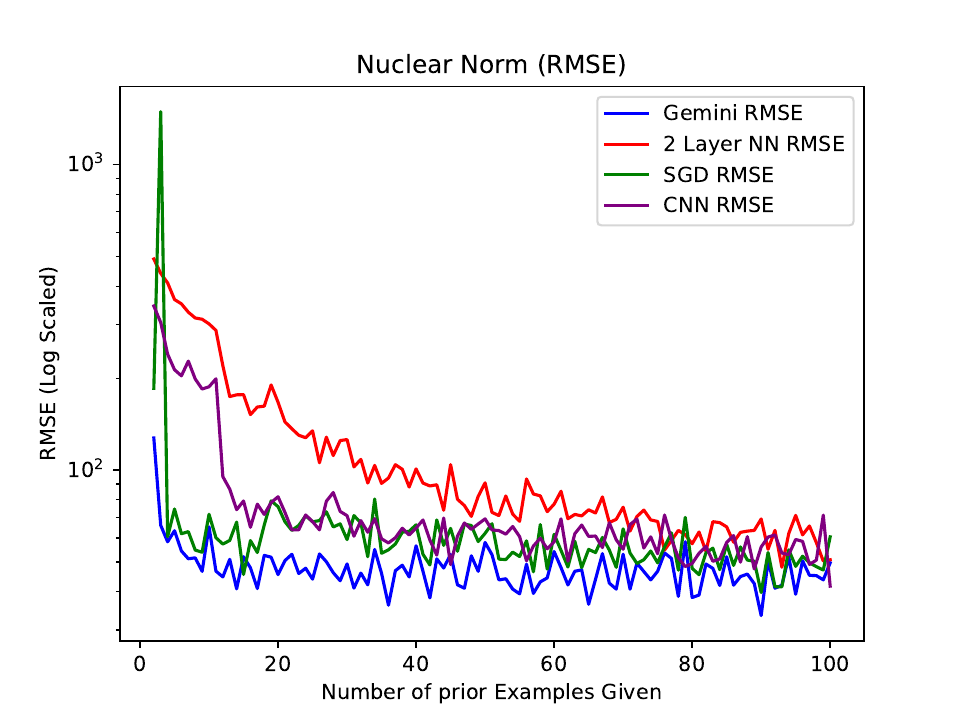}
            % Table for Nuclear Norm
            \vspace{0.5cm}
            \begin{tabular}{|l|c|}
                \hline
                \textbf{Model} & \textbf{Average RMSE (Last 25 Trials)} \\
                \hline
                Gemini & 45.517405 \\
                2 Layer NN & 60.635421 \\
                SGD & 51.442399 \\
                CNN & 55.336333 \\
                \hline
            \end{tabular}
        \end{minipage}
    }
    \hspace{1 cm} % Space between the two subfigures
    \subfigure[RMSE of the top-3 singular values for every model in the SVD learning task over all 100 prior examples. The table below shows the average RMSE of the top-3 singular values over the last 25 trials. \label{fig:rmse_svd_all_models}]{
        \begin{minipage}{0.45\textwidth}
            \centering
            \includegraphics[width=\textwidth, keepaspectratio]{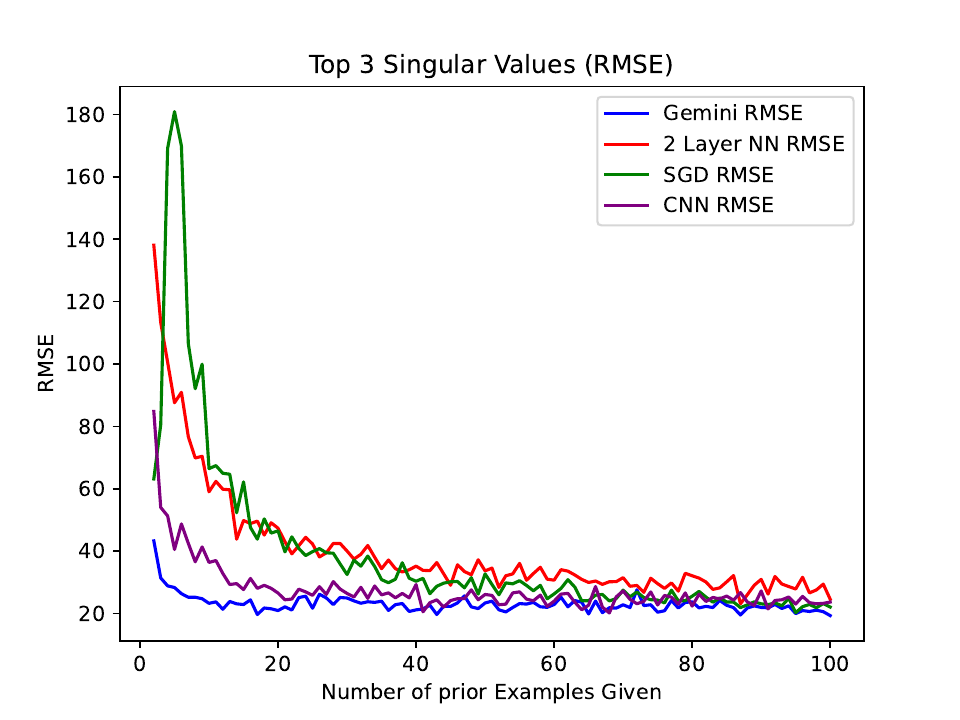}
            % Table for SVD
            \vspace{0.5cm}
            \begin{tabular}{|l|c|}
                \hline
                \textbf{Model} & \textbf{Average RMSE (Last 25 Trials)} \\
                \hline
                Gemini & 21.788701 \\
                2 Layer NN & 28.871207 \\
                SGD & 23.579583 \\
                CNN & 24.372482 \\
                \hline
            \end{tabular}
        \end{minipage}
    }
    \caption{RMSE for the nuclear norm and SVD learning tasks when given 5x5 matrices as inputs.}
\end{figure*}

\begin{figure*}[!ht]
    \centering
    \includegraphics[width=0.9\textwidth, keepaspectratio]{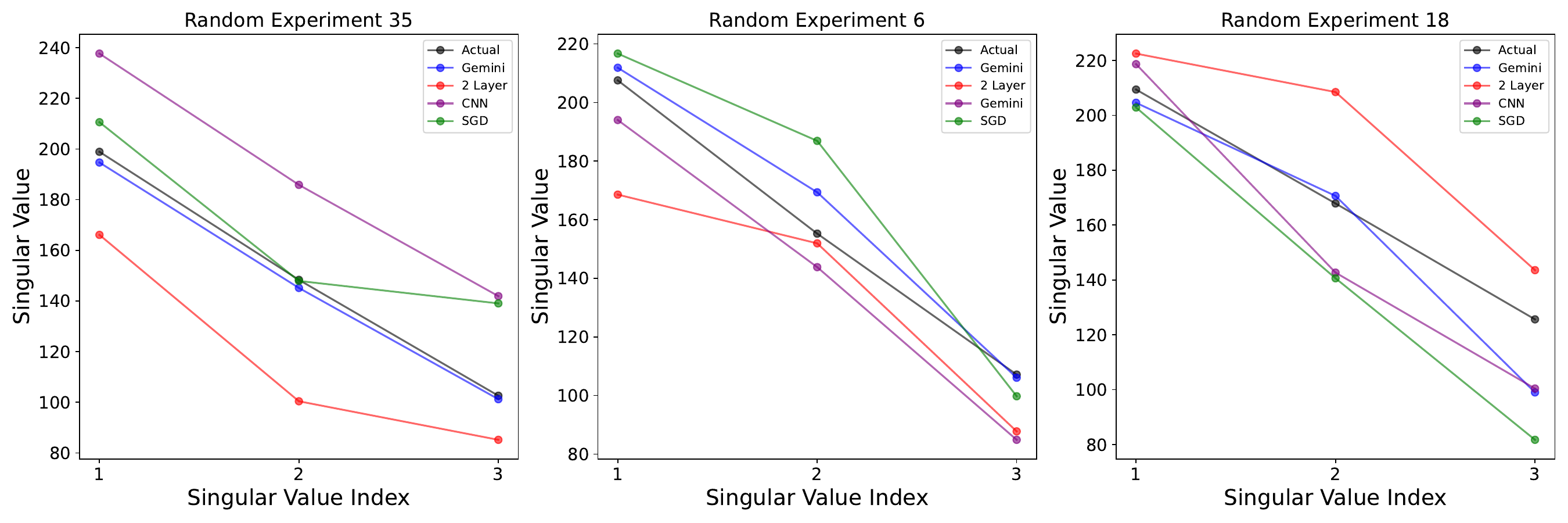}
    \caption{Average predictions compared to average actual values for the SVD learning task.}
    \label{fig:svd_average_predict}
\end{figure*}

\begin{figure*}[!ht]
    \centering
    \subfigure[Plot of RMSE of the top-3 singular values for each model on 10x10 matrices over all 50 trials. The table below shows the average RMSE for trials 35 to 50. \label{fig:rmse_svd_10x10}]{
        \begin{minipage}{0.45\textwidth}
            \centering
            \includegraphics[width=\textwidth, keepaspectratio]{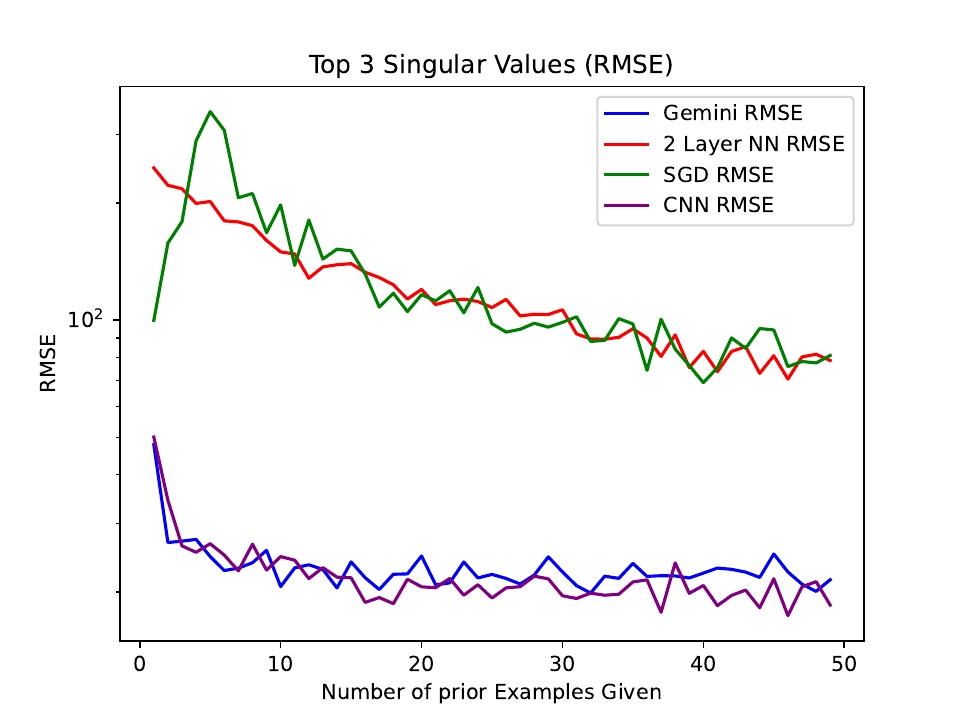}
            % Table for 10x10 matrix
            \vspace{0.5cm}
            \begin{tabular}{|l|c|}
                \hline
                \textbf{Model} & \textbf{Average RMSE (Trials 35 to 50)} \\
                \hline
                Gemini & 22.317223 \\
                2 Layer NN & 81.395695 \\
                SGD & 82.747149 \\
                CNN & 19.984936 \\
                \hline
            \end{tabular}
        \end{minipage}
    }
    \hspace{1 cm} % Space between the two subfigures
    \subfigure[RMSE of the top-3 singular values for each model on 25x25 matrices over all 50 trials. The table below shows the average RMSE for trials 35 to 50. \label{fig:rmse_svd_25x25}]{
        \begin{minipage}{0.45\textwidth}
            \centering
            \includegraphics[width=\textwidth, keepaspectratio]{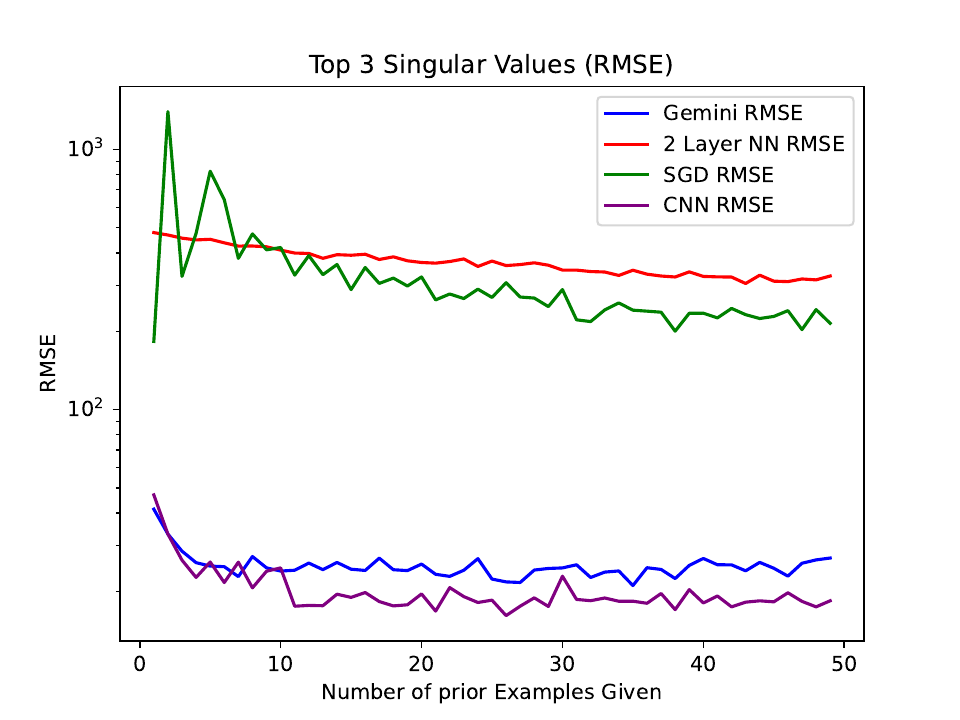}
            % Table for 25x25 matrix
            \vspace{0.5cm}
            \begin{tabular}{|l|c|}
                \hline
                \textbf{Model} & \textbf{Average RMSE (Trials 35 to 50)} \\
                \hline
                Gemini & 24.234574 \\
                2 Layer NN & 334.379760 \\
                SGD & 241.310389 \\
                CNN & 18.497619 \\
                \hline
            \end{tabular}
        \end{minipage}
    }
    \caption{RMSE for SVD learning tasks when given 10x10 and 25x25 matrices as inputs.}
\end{figure*}

\begin{figure*}[!ht]
    \centering
    \includegraphics[width=0.45\textwidth, keepaspectratio]{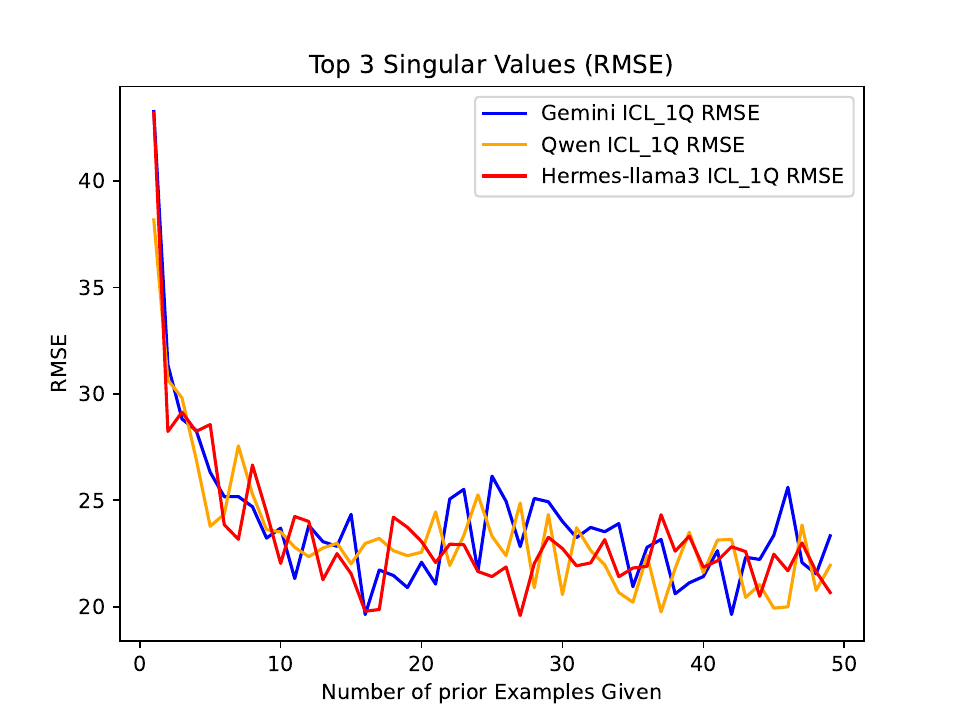}
    \caption{RMSE of the top-3 singular values for Open Source Models in SVD learning task on 5x5 matrices over all 50 trials.}
    \label{fig:rmse_svd_open_source}
\end{figure*}

%\begin{figure}[!ht]
%    \centering
%    \vspace{-1em}
%    \includegraphics[width=0.8\textwidth, keepaspectratio]{Styles/Nuclear-Norm_Plots/RMSE_Nuclear_Norm_All_Models.pdf}
%    \caption{RMSE of the nuclear norm predictions for every model in the nuclear norm learning task.}
%    \label{fig:rmse_nuclear_norm_all_models}
%
%\end{figure}

%\begin{figure}[!ht]
%    \centering
%    \includegraphics[width=0.8\textwidth, keepaspectratio]{Styles/SVD_Plots/RMSE_SVD_All_Models.pdf}
%    \caption{RMSE of the top 3 singular values for every model in the SVD learning task.}
%    \label{fig:rmse_svd_all_models}
%\end{figure}

%Figure  
%(\autoref{fig:svd_average_predict} and
%\autoref{fig:rmse_svd_all_models}) %illustrates the performance of different models in predicting the top 3 singular values of a 5x5 matrix.

Figure \ref{fig:rmse_svd_all_models} shows the Root Mean Square Error (RMSE) for each model in predicting the top-3 singular values of a 5x5 matrix, as a function of the number of prior examples used. As observed in the nuclear norm results, the SGD Linear Regression model performs competitively with fewer examples but again exhibits significant error spikes as more examples are introduced, likely due to overfitting. In contrast, Gemini maintains stable performance without signs of overfitting, even with more training examples. Gemini also consistently outperforms both the 2-layer NN and CNN models.

Figure \ref{fig:svd_average_predict} presents multiple subplots (1a - 1c) representing random experiments selected from 3 randomly selected experiments across all models. In each experiment, we evaluate how well the models predict all 3 singular values. Gemini shows the most accurate predictions, consistently outperforming other models in all three subplots. While some models may predict one singular value more accurately in specific cases, Gemini demonstrates superior performance across all 3 values overall.

These results suggest that Gemini offers both superior accuracy and consistency compared to classical models when predicting the top-3 singular values of a matrix. Just like with the nuclear norm—a more complex task than the $p$-norms—Gemini demonstrates clear advantages here as well. Notably, even with fewer training examples, Gemini achieves performance comparable to that of a CNN trained with significantly more examples. Even as the number of prior examples increases, Gemini continues to maintain stable performance and avoids overfitting, while classical models like SGD Linear Regression struggle. This reinforces the trend we have observed throughout our analysis: the more complex the task, the more Gemini outperforms traditional models.

\subsection{Scalability With Larger Matrices}

Figure \ref{fig:rmse_svd_10x10} shows the Root Mean Square Error (RMSE) for Gemini, a 2-Layer Neural Network (NN), and a Convolutional Neural Network (CNN) when tasked with finding the top-3 singular values of a 10x10 matrix. Similarly, Figure \ref{fig:rmse_svd_25x25} presents the RMSE results for the same models when applied to a 25x25 matrix.

As these results demonstrate, the performance of In-Context Learning (ICL) for finding the top-3 singular values scales with the size of the matrix. Notably, the 2-layer NN and SGD Linear Regression begin to show signs of deteriorating accuracy, particularly in the 25x25 case, as compared to their performance on 5x5 matrices. In contrast, Gemini continues to deliver competitive results, even when faced with larger, more complex matrices.

Interestingly, Gemini's performance remains comparable to that of the CNN, a model traditionally better suited for handling high-dimensional data. This observation suggests that LLMs, through ICL, can serve as an efficient and effective alternative to conventional models like the 2-layer NN, particularly when data availability is limited. LLMs can potentially learn from fewer prior examples and still achieve similar accuracy levels.

Unlike CNNs, 2-layer NNs, and SGD Linear Regression which require extensive training to optimize their parameters, LLMs using ICL can generate accurate predictions entirely through inference, without the need for explicit training on the specific task. This presents a significant advantage, as it allows LLMs to quickly produce results based solely on observed examples, making them a highly flexible solution for small datasets or cases where training resources are limited. Even for 25x25 matrices, Gemini's performance remains within a competitive range of the CNN, with only a marginal difference in RMSE, highlighting the potential of LLMs for efficiently tackling high-dimensional tasks.

\subsection{Comparisons with Open Source Models}
Figure \ref{fig:rmse_svd_open_source} shows the Root Mean Square Error (RMSE) for Gemini, Qwen, and Hermes Llama3 when tasked with finding the top-3 singular values in the SVD learning task. The results show that not only can advanced models like Gemini use In-Context Learning (ICL) to perform complex matrix computations, but smaller open-source models, such as Qwen and Hermes Llama3, can also demonstrate strong performance using ICL. This suggests that the ability to learn and execute tasks in context is not exclusive to large, proprietary models but can be extended to more lightweight, portable, and publicly available models as well.

There are many implications of this. First, it reveals to us that ICL is not reserved to state-of-the-art models but is a broader phenomenon observed across many language models. This expands the reach of ICL, making it accessible for use in various applications, even in environments where computational resources or access to large models like Gemini may be limited. Second, the use of open-source models offers a valuable opportunity for deeper exploration into why ICL even occurs. Since these models are more accessible and often easier to modify, they provide a window into understanding the underlying principles of ICL. By experimenting with and analyzing the behavior of open-source models, researchers can potentially unlock new insights into how ICL operates and why it emerges in models of varying complexity.

This opens up new ways for understanding how smaller models can generalize complex tasks in context, and for exploring the cause of ICL across different LLMs. As open-source models demonstrate similar capabilities to larger models, they may become pivotal in future research.

\section{Conclusions}

In this paper, we provided preliminary proof-of-concept results demonstrating that In-Context Learning (ICL) in Large Language Models (LLMs) shows highly promising results, particularly when tasked with solving complex problems, such as computing the top-k singular values of a matrix with minimal prior examples.

Our experiments revealed several interesting properties of LLMs. While they performed well on simpler tasks, such as computing vector p-norms—outperforming the classic Stochastic Gradient Descent (SGD) Linear Regression algorithm—they were still outperformed by a standard 2-layer Neural Network (NN). However, as the complexity of the tasks increased, particularly with the nuclear norm and top-k singular values of a matrix, the LLM's strengths became more apparent. When compared to SGD-based Linear Regression, a 2-layer NN, and even a Convolutional Neural Network (CNN), the LLM consistently outperformed these models. Unlike CNNs and 2-layer NNs, which require extensive training to optimize their parameters, LLMs using ICL generated accurate predictions entirely through inference, without the need for task-specific training. Remarkably, the LLM achieved these results with very few prior examples, demonstrating rapid error convergence and maintaining high performance, even as the complexity and size of the matrix increased.

Additionally, we found that the LLMs scaled well with varying matrix sizes, performing comparably on larger matrices like 10x10 and 25x25, which traditionally challenge more conventional models. Open-source models, such as Qwen and Hermes Llama3, also showed promising results, indicating that ICL's potential is not limited to proprietary models like Gemini. These findings further support the idea that ICL could provide a scalable and efficient alternative to traditional methods in high-dimensional problems.

In summary, we found that In-Context Learning offers a promising new approach to solving non-linear regression tasks. As problems grow in complexity, ICL could provide a highly efficient and accurate method for addressing these challenges, potentially transforming how we tackle large-scale computations.

\section*{Acknowledgments}
Research was sponsored by the Army Research Office and was accomplished under Grant Number W911NF-24-1-0397. The views and conclusions contained in this document are those of the authors and should not be interpreted as representing the official policies, either expressed or implied, of the Army Research Office or the U.S. Government. The U.S. Government is authorized to reproduce and distribute reprints for Government purposes notwithstanding any copyright notation herein.

%\clearpage
\

\bibliographystyle{ieeetr}

\end{document}